# BOISHOMMO: Holistic Approach for Bangla Hate Speech

Md Abdullah Al Kafi

Department of Computer Science and Engineering

Daffodil International University, Dhaka, Bangladesh

alkafi.1408@gmail.com

Sumit Kumar Banshal

Department of Computer Science and Engineering

Alliance University, Bangalore-562106, India

**Corresponding Author:** sumitbanshal06@gmail.com

Md Sadman Shakib

Department of Computer Science and Engineering

Daffodil International University, Dhaka, Bangladesh

sadmanshakib345@gmail.com

Showrov Azam

Department of Computer Science and Engineering

Daffodil International University, Dhaka, Bangladesh

showrovazam230@gmail.com

Tamanna Alam Tabashom

Department of Computer Science and Engineering

Daffodil International University, Dhaka, Bangladesh

tamamnaalamtabashom0202@gmail.com

April 11, 2025

**Abstract**

One of the most alarming issues in digital society is hate speech (HS) on social media. The severity is so high that researchers across the globe are captivated by this domain. A notable amount of work has been conducted to address the identification and alarm system. However, a noticeable gap exists, especially for low-resource languages. Comprehensive datasets are the main problem among the constrained resource languages, such as Bangla. Interestingly, hate speech or any particular speech has no single dimensionality. Similarly, the hate component can simultaneously have multiple abusive attributes, which seems to be missed in the existing datasets. Thus, a multi-label Bangla hate speech dataset named "BOISHOMMO" has been compiled and evaluated in this work. That includes categories of HS across race, gender, religion, politics, and more. With over two thousand annotated examples, BOISHOMMO provides a nuanced understanding of hate speech in Bangla and highlights the complexities of processing non-Latin scripts. Apart from evaluating with multiple algorithmic approaches, it also highlights the complexities of processing Bangla text and assesses model performance. This unique multi-label approach enriches future hate speech detection and analysis studies for low-resource languages by providing a more nuanced, diverse dataset.9

# 1 Introduction

Hate speech (HS) is a form of expression that spreads negativity and might incite violence against people or groups based on their inborn traits, including Religion, Race, Nationality, Sexual Orientation, and Gender. The increasing prevalence of social media platforms has contributed to spread over a broader section of society. The nature of communication provided over social media has extended beyond an individual's psychological traits and backgrounds [Maruf et al., 2024]. Out of 8.08 billion people living on the earth, more than 5 billion are using social media[1]. Hate speech in online communities polarises individuals and incites disorder, occasionally resulting in real-world violence and conflict. It can disseminate misinformation, undermining democracy and inciting violence. Texting and chatting serve as methods of communication on social media. When text is employed to disparage others, it is deemed erroneous and defamatory, potentially resulting in harassment, religious discord, and eve-teasing on social media platforms [Mridha et al., 2021]. Moreover, hate speech can inflict significant psychological harm on its victims, resulting in depression and anxiety disorders. Conversely, hate speech may simultaneously target various aspects. Identifying and addressing hate speech is crucial for fostering a secure online environment where individuals can engage without fear of harassment or discrimination. Numerous nations have implemented stringent legislation prohibiting hate speech. A comprehensive automated system that effectively identifies hate speech would mitigate its impact on social space. Natural Language Processing (NLP) is a significant subfield of artificial intelligence that enables computers to comprehend, translate, and manipulate human language. [Lucky et al., 2021]. Natural Language Processing enables individuals to develop automated instruments for managing language-related tasks. Such tools necessitate the utilization of an annotated dataset that encapsulates the intricacies and diversities of hate speech. Annotated datasets and linguistic resources are fundamental to natural language processing (NLP) and machine learning.

---

[1]https://wearesocial.com/uk/blog/2024/01/digital-2024-5-billion-social-media-users/

Bangla ranks as the seventh most spoken language, with 265 million speakers, both native and non-native, globally [Sharma and Chatterjee, 2021], serving as a medium for communication and emotional expression across various cultures. Nonetheless, it is classified as a low-resource language, akin to Assam anda (India), Sinhala (Sri Lanka), and Nepali (Nepal), owing to the scarcity of quality digital data. The Bangla language possesses distinctive linguistic characteristics, setting it apart from other low-resource languages. Bangla is recognized for its intricate grammatical framework, diverse vocabulary, and distinctive syntax. Bangla is a phonetic language, as it assigns a distinct letter to each sound, facilitating simpler spelling for individuals; however, this characteristic introduces greater complexity in machine processing[Akan et al., 2023]. The Bangla language is profoundly linked to the history, literature, and arts of the Bangla people, in addition to its cultural significance. Bangla literature and music profoundly influence the cultural landscape of South Asia.

The rising prevalence of social media in regions such as Bangladesh and West Bengal necessitates the development of efficient hate speech detection mechanisms in the Bangla language. Despite significant advancements in Natural Language Processing (NLP) research and tool development for the Bangla language in recent years, it remains inferior to high-resource languages such as English, particularly in the domain of hate speech detection, where hate speech can simultaneously target multiple attributes and encom-pass various categories—current datasets for hate speech detection in the Bangla language lack multi-labeled datasets.

This research significantly advances the field in multiple critical aspects. Initially, it offers an improvised and comprehensive annotated multi-label dataset, "Boishommo," which encapsulates the varied and frequently nuanced manifestations of hate speech in Bangla. This dataset establishes a new standard for hate speech detection in low-resource and complex languages, rendering it a valuable and dependable resource for future research. This study improves the ability of current models to process and comprehend Bangla text more efficiently. The extensive ramifications of this research are substantial. This study enhances hate speech detection in Bangla, fostering safer and more inclusive online environments for Bangla-speaking communities. It further promotes the overarching objective of cultivating more respectful global digital environments, particularly in settings with constrained linguistic resources.

## 2 Literature Review

Hate speech encompasses any mode of communication, whether verbal, written, or behavioral, that conveys animosity or violence towards an individual or group based on race, religion, or other characteristics. It may include abusive, discriminatory, or hostile language or expressions. Hate speech can manifest in various forms, including verbal, non-verbal, metaphorical, and symbolic expressions, among others. Bullying, scapegoating, or stigmatization may constitute hate speech [Recupero et al., 2022].

Hate speech may vary based on the modes of expression employed. Hate speech disseminated online or through social media can exert a unique influence on society, in contrast to hate speech expressed in public venues [Chetty and Alathur, 2018] . Individuals who experience hate speech or similar conduct in real life frequently suffer from psychological trauma, social avoidance, and a diminished quality of life. Occasionally, they are compelled to relocate residences. Facebook's protected characteristics delineate hate speech as an assault on an individual's dignity, encompassing their race, origin, or ethnicity. Twitter policies stipulate that tweets must not be utilized to threaten or harass individuals based on their ethnicity, gender, religion, or any other characteristic. Besides age, caste, and disability, YouTube also restricts content that incites violence or hatred against specific individuals or groups [Alkomah and Ma, 2022].

Nonetheless, the comment section is not the sole platform for disseminating animosity; users also propagate discriminatory content against social groups based on attributes such as religion, sexual orientation, gender, or disabilities, a phenomenon known as hate speech. [Erjavec and Kovačič, 2012]. Prior research indicates that user comments function as exemplars that influence the perception of public opinion and impact the attitudes of those who read them.[Peter et al., 2014, Neubaum and Krämer, 2017, Hsueh et al., 2015]

Online hate speech on social media can lead to various effects, including self-doubt, depression, and insecurity, which are the most prevalent [Vedeler et al., 2019]. Recognizing and denouncing individuals associated with hate speech in public settings is comparatively easier than on social media platforms. This renders it one of the most critical subjects for regulation on social media. Owing to its significance and influence, hate speech detection is a prominent research domain in natural language processing and computational linguistics. Offensive remarks, including hate speech and cyberbullying, have been among the most extensively studied topics in Natural Language Processing (NLP) over the past few decades. [Pestian et al., 2010, Pillay and Solorio, 2010, León-Paredes et al., 2019, Moreno et al., 2019, Wang et al., 2021].

Text classification has been extensively researched and employed in numerous real-world applications, including detecting hate speech, over the past few decades. The recent advancements in NLP and text mining have piqued the interest of many researchers in developing applications that utilize text classification methods. [Mullah and Zainon, 2021]. Additionally, researchers conducted numerous experiments and created valuable tools and techniques to generate and process Bangla language materials. [Sen et al., 2022]. Scholars from various disciplines have conducted numerous studies on detecting hate speech in the past decade. Classical machine learning and contemporary deep learning methodologies are implemented in these investigations. The hate speech detection problem is typically presented as a text classification task where raw text data constitutes the pipeline's base input.

Support Vector Machine (SVM) has been found to be better suited among machine learning approaches for detecting hate speech from social media datasets [Mercan et al., 2021]. Even after implementing different feature engineering approaches on the dataset, the SVM outperformed other traditional approaches of machine learning [Abro et al., 2020]. The efficacy of conventional machine learning techniques for hate speech detection has been evaluated against various other studies, concluding that Support Vector Machine, Random Forest, and KNN are the predominant algorithms when utilized alongside the TF-IDF vectorization method citepMullah2021. An ensemble method employed to detect hate speech from a South African tweet dataset, the ensemble approach integrated Random Forest, Support Vector Machine, Gradient Boosting, and Logistic Regression [Oriola and Kotze, 2020]. These methodologies are prominently used to validate the datasets as well.

Moreover, hate speech can be divided into categories, making it a multi-class or multi-label problem [Khan et al., 2021]. This phenomenon was discussed in implementing different studies over the period [Modha et al., 2020, Mishra et al., 2021, Chiril et al., 2022, Banshal et al., 2023]. However, multi-class classification often misses the inclusivity of different classes in a single instance, which can be addressed by multi-label classification [Hristozov et al., 2008, Cheng and Hüllermeier, 2009, Mishra et al., 2021]. Also, techniques such as binary relevance and classifier chain are most common in multi-label classification in traditional Machine learning workflows [Hristozov et al., 2008, Cheng and Hüllermeier, 2009, Mishra et al., 2021].

Besides class inclusivity, context and language are pivotal in detecting and classifying hate speech. Numerous studies have investigated hate speech in diverse languages and cultural contexts. To explore the contextual usage of hate speech, several country-wise contexts have been evaluated, such as Hindi and Bangla within the Indian context [Ghosal and Jain, 2023], English in the South African context [Oriola and Kotze, 2020], Indonesian context [**?**], Urdu in Pakistani context [Arshad et al., 2023]. Nonetheless, substantial opportunities for research remain in multi-label hate speech detection in Bangla, a complex language presenting distinctive challenges that have not been thoroughly addressed. The scarcity of labeled datasets for Bangla, and therefore, the usage of distinct datasets, kept this domain unnurtured in broader aspects [Maruf et al., 2024].

Nonetheless, no dependable multi-label hate speech dataset for the Bangla language has been identified. English is the primary language for online resources, technical knowledge, journals, and documentation. As a result, numerous Bangla-speaking individuals with limited proficiency in English encounter obstacles in accessing English resources. Moreover, linguistic com-plexity is a diverse concept; limited resources exacerbate the difficulty. To comprehend its complex nature, one must examine various attributes, including phonology, morphology, syntax, semantics, and grammar [Comrie, 1988]. . Bangla is a complex and low-resource language due to its limited digital presence and derivation from ancient Sanskrit, considered one of the most intricate languages. [Encyclopaedia Britannica, 2024]. Bangla still inherits some of its ancestor's complexities [Chowdhury et al., 2021, Islam, 2023]. These complexities need to be addressed in a multi-label hate speech context.

A multi-label Bangla social media hate speech dataset is essential for addressing the gaps above. This study seeks to create a dataset and compare machine learning algorithms to establish baseline performance metrics. This will establish a basis for subsequent research in this domain.

## 3 Corpus Development

Proper data collection and annotation are very important for creating a good dataset. Tasks like hate speech annotation on inclusive categories add more complexity to it. This paper follows predefined steps in the dataset creation process to address this complexity. To generate a multifaceted dataset, ten broader categories of hate speech were identified [Fortuna and Nunes, 2018]. Each category indicates specific forms of discriminatory language and harmful discourse frequently encountered on social media platforms. The data points were collected from Facebook [Facebook, 2024], a widely used social networking platform. The official pages of popular newspapers, including Prothomalo, Juganttor, and Kaler Kontho, have been utilized as data sources to reach a broader audience. The data collection has been concentrated on the most frequently commented news from the 2020-21 year [Prothomalo, 2024, Jugantor, 2024, Kontha, 2024]. [2]

The selection was made with the primary objective of identifying offensive words and hate speech pertinent to the Bangla-speaking community to capture a diverse range of user interactions. This process has resulted in compiling and processing 2499 comments for further analysis. Three annotators validated each comment, and the inter-annotator agreements were verified. Depending on the majority vote, the final dataset was developed. Further, the dataset was evaluated using popular machine learning algorithms. The procedural elaboration of annotation and dataset compilation is illustrated in the following parts. Figure 1 shows the steps in the dataset creation process, where all the steps are shown as a workflow diagram.

---

[2]Facebook: https://www.facebook.com/
Prothom Alo: https://www.facebook.com/DailyProthomAlo
Kaler Kantho: https://www.facebook.com/kalerkantho
Jugantor: https://www.facebook.com/DainikJugantor

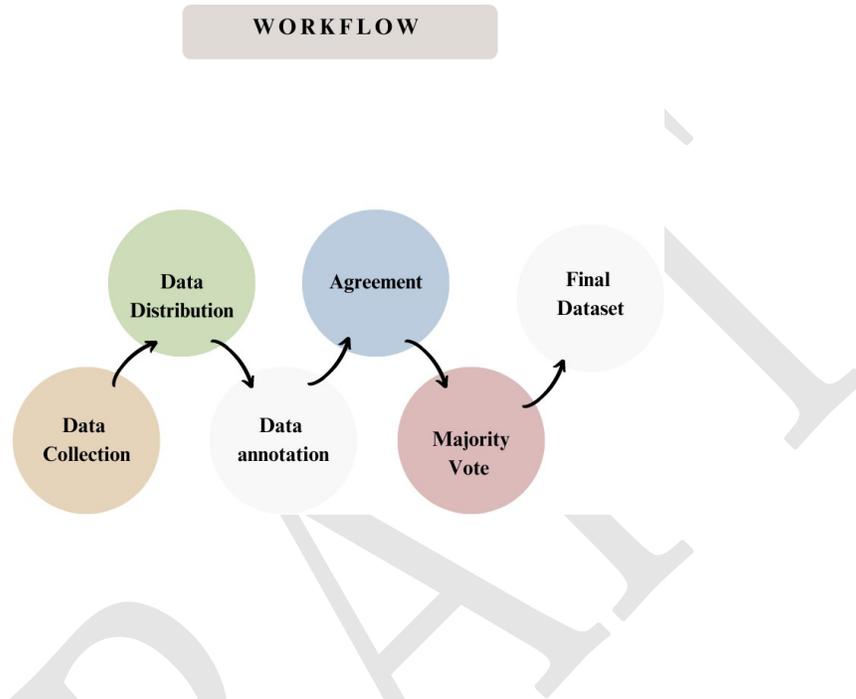

Figure 1: Corpus Development Workflow Diagram

## 3.1 Data Annotation

Each of the comments was annotated by three annotators, which resulted in a 2499*4 matrix comprising three labels and comments. All annotators who are native Bangla speakers and have completed at least two years of undergraduate studies were selected. These two conditions were meticulously conveyed to ensure the morphological comprehension of the language and the capacity to discern the words' concealed meaning or fundamental purpose. The consistency of labeling was annotated comments was ensured by using the following process:

- Cohen's Kappa was calculated to measure inter-annotator agreement across pairs of annotators (Annotator 1 vs. Annotator 2, Annotator 1 vs. Annotator 3, Annotator 2 vs. Annotator 3). This ensured that the labeling adhered to the agreed guidelines.

- Any inconsistencies in labeling were resolved through a consensus-based approach.

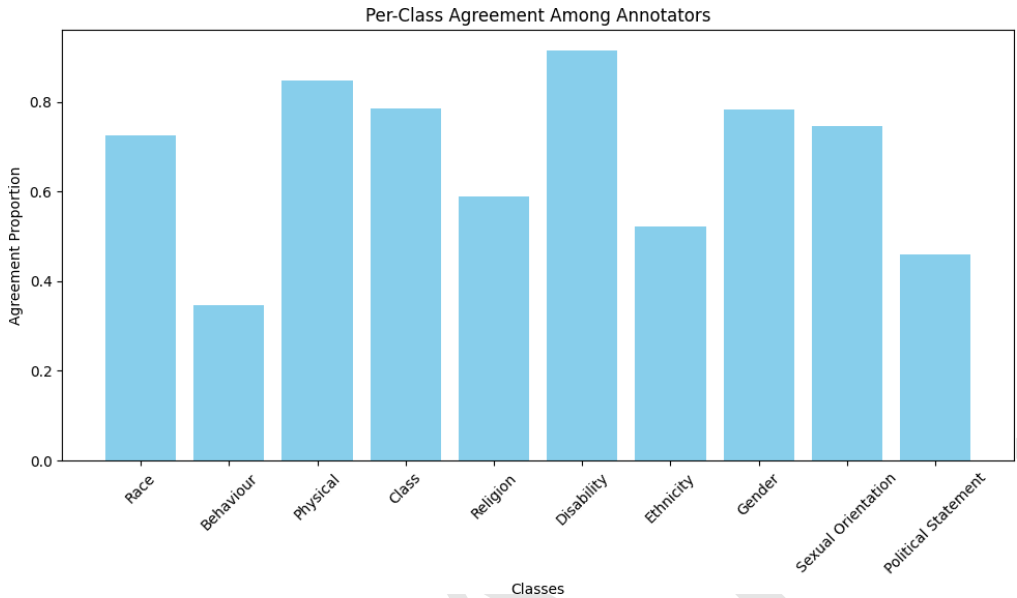

Figure 2: Inter Annotator Agreement Value Distribution over Categories

For annotations, majority voting was followed, which means that for each class label, the majority vote was taken (i.e., if two or more annotators labeled the instance as 1, the final label is 1). This resulted in a new dataset, which represents the consensus among annotators.

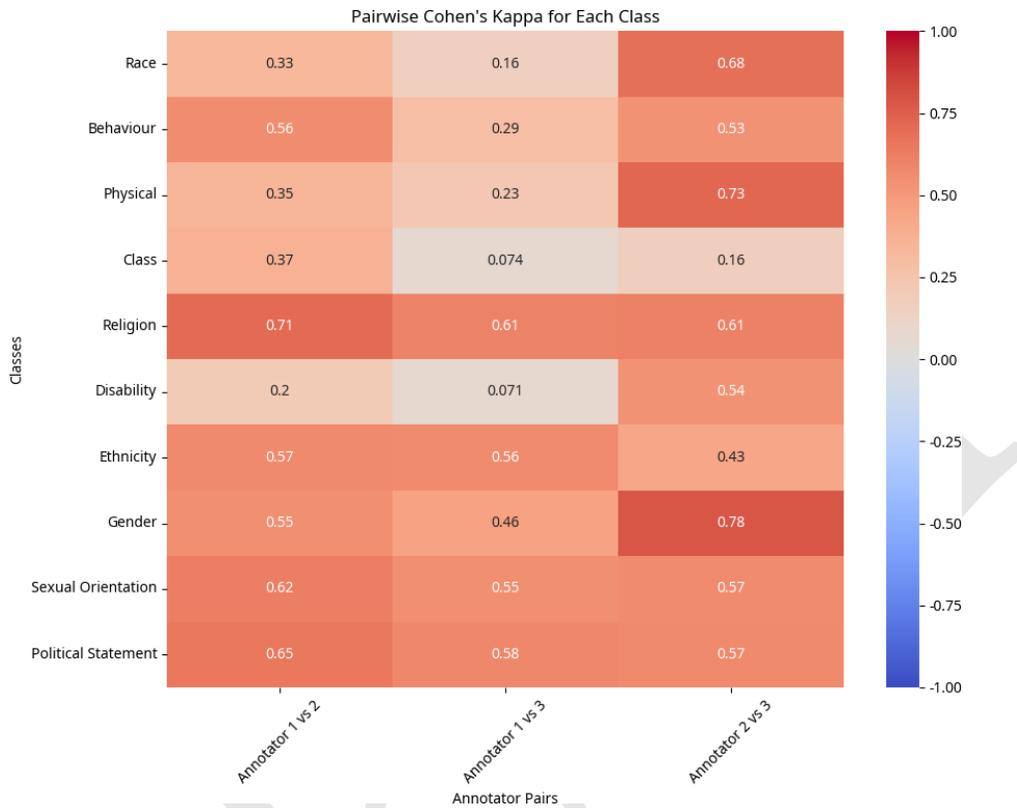

Figure 3: Inter Annotator Agreement (Cohen's Kappa) Heat Map

Figure 2 and Figure 3 show pairwise Cohen's Kappa values for each of the ten categories (Race, Behavior, Physical, Religion, Class, Nationality, Gender, Sexual Orientation, and Political Statement) between our three annotators (Annotator 1, Annotator 2, and Annotator 3). The heatmap measures inter-annotator agreement highlighted by a color gradient corresponding to Kappa values, showing a range from -1 to +1, where the higher value indicates better agreement between the annotators.

### 3.1.1 Category-wise Analysis

The kappa value refers to Cohen's kappa, a statical measurement to assess the level of agreement between two or more annotators [Cohen, 1960].

- **Calculation:** The kappa value is calculated using the observed agreement between raters and the expected agreement. The formula is:

$$\kappa = \frac{P_o - P_e}{1 - P_e}$$

Where:

- $P_o$ = observed agreement
- $P_e$ = expected agreement by chance.

- **Value Range:** Kappa values range from -1 to 1:

    - $\kappa < 0$: Indicates less agreement than would be expected by chance.
    - $\kappa = 0$: Indicates no agreement beyond chance.
    - $0 < \kappa < 1$: Indicates varying degrees of agreement.
    - $\kappa = 1$: Perfect agreement.

- **Interpretation:** Common interpretation guidelines (though they can vary) are:

    - $\kappa < 0$: Poor agreement
    - $0 \leq \kappa < 0.20$: Slight agreement
    - $0.21 \leq \kappa < 0.40$: Fair agreement
    - $0.41 \leq \kappa < 0.60$: Moderate agreement
    - $0.61 \leq \kappa < 0.80$: Substantial agreement
    - $0.81 \leq \kappa < 1.00$: Almost perfect agreement

For this study, the following Kappa values for each category were achieved and are shown in Figure 3 and Figure 2.

1. **Race**:

    - Kappa: Annotator 1 vs 2 = 0.33 (fair), 1 vs 3 = 0.16 (slight), 2 vs 3 = 0.68 (substantial).
    - Discussion: Highest agreement between Annotators 2 and 3; lowest between 1 and 3, indicating potential subjectivity.

2. **Behavior**:
    - Kappa: 1 vs 2 = 0.56 (moderate), 1 vs 3 = 0.29 (fair), 2 vs 3 = 0.53 (moderate).
    - Discussion: Moderate to fair agreements suggest this category was easier to annotate than race.

3. **Religion**:
    - Kappa: 1 vs 2 = 0.71, 1 vs 3 = 0.61, 2 vs 3 = 0.61 (all substantial).
    - Discussion: High agreement across all annotators, indicating clearer interpretation.

4. **Class**:
    - Kappa: 1 vs 2 = 0.37 (fair), 1 vs 3 = 0.074 (slight), 2 vs 3 = 0.16 (slight).
    - Discussion: Low agreement, especially involving Annotator 3.

5. **Nationality**:
    - Kappa: 1 vs 2 = 0.57, 1 vs 3 = 0.56, 2 vs 3 = 0.43 (all moderate).
    - Discussion: Moderate agreement across all pairs, showing easier annotation.

6. **Gender**:
    - Kappa: 1 vs 2 = 0.55, 1 vs 3 = 0.46 (both moderate), 2 vs 3 = 0.78 (substantial).
    - Discussion: Strong agreement between 2 and 3; slightly lower with Annotator 1.

7. **Sexual Orientation**:
    - Kappa: 1 vs 2 = 0.62 (substantial), 1 vs 3 = 0.55, 2 vs 3 = 0.57 (both moderate).
    - Discussion: Consistent agreement between annotators.

8. **Political**:

- Kappa: 1 vs 2 = 0.65 (substantial), 1 vs 3 = 0.58, 2 vs 3 = 0.57 (both moderate).
- Discussion: Moderate to substantial agreement, showing solid consistency across annotators.

Here, categories such as Religion, Sexual Orientation, and Politics tend to show higher agreement between most annotator pairs. In contrast, a slightly lower agreement can be seen in categories like race and class.

## 3.2 Dataset Compilation

In order to preserve the integrity of the annotation process, the collected comments were dispersed across various sets of files. Individual annotators were provided with the segregated files to validate the comments and classified them into ten categories: Race, Behaviour, Physical, Class, Religion, Disability, Nationality, Gender, Sexual Orientation, and Political. All comments were combined into a single matrix for subsequent processing after the annotations were compiled. Binary value-based labeling was implemented for all categories assigned either 0 or 1. Enabling the simultaneous labeling of each comment into multiple categories. A majority voting mechanism was implemented throughout the annotated dataset to generate a final, dependable set of annotations. This approach generated a consensus-based dataset, with the majority label for each chosen instance. The outcome was a refined dataset that accurately represented the consensus of the annotators, thereby resolving inconsistencies in the original annotations. The multi-labeled annotated dataset, 'BOISHOMMO,' was corroborated by this meticulously controlled and structured approach.

Figure 4 shows a noteworthy feature of the dataset lies in the carefully crafted alignment of comments, each belonging to multiple labels. This structural alignment is evident throughout the dataset, with each comment consistently labeled across all categories by all the annotators.

Figure 5 shows the top 5 most frequent words in the dataset. The word with the highest frequency is on the left, and the words with lower frequency are on the right side. Here the most occurring word is "অতয়5", appearing over 400 times, followed by "মালেয়িশয়া", "তথয়", "িবজ্ঞিঃ", and "এক" ; where "এক" has the lowest occurring time.

Figure 6 represents the visual proportional breakdown of the percentage of existence in each category of hate speech. Here, the number of hate speech related to Behaviour, Nationality, Religion, Political Statement, and Race are higher, respectively, 22.2%, 13.3%, 16.3%, 10.6%, and 10.4%. Although the categories are mutually inclusive, the proportion of each category supports previously discussed Kappa values from Figure 3.

| No | Sentence | Category |
|----|----------|----------|
| 1 | সর্বসমক্ষে আসামিদের পরিবার পরিজনদের সামনে ফাঁসি দেওয়া উচিত! যে চলে গেছে সে আর ফিরবে না! কিন্তু আসামিদের তিল তিল করে দগ্ধে মারা উচিত! | Physical, Class |
| 2 | যে শালা মালাউন বলেছে ও বোনের চুদি তোমাদের মুখে দাড়ি মাথায় টুপি রেখে ভন্ডামী কর তোমরা জংগী? | Behavior, Class, Religion |
| 3 | বিরাট কুত্তার বাচ্চাটা ব্যাটসম্যান হিসেবে যতোটা ভাল চরিত্রটা ততোটাই খারাপ। এই কুত্তার বাচ্চার মায়েরে মনে হয় রেডিয়ার কতগুলা নেড়িকুত্তা গণধর্ষন করে বীর্য দিয়ে ভরে দিয়েছিল। তারপর কুত্তার বাচ্চার মা এইটারে জন্ম দিয়েছে। | Race, Behaviour, Sexual Orientation |
| 4 | কি বলবো বুঝতেছিনা! যানি ওরা অসহায় তারপরেও এমন কাজ করা ঠিক না ওরা চাইলে কাজ করে খেতে পারে! | Physical, Disability |
| 5 | পুরো বাংলাদেশে আওয়ামীলীগ সন্ত্রাসীরা,, শতশত একর জমি,, জোর করে দখল করে নিয়েছে,, দীর্ঘদিন জোর করে ক্ষমতায় থাকার ফসল এই সব সন্ত্রাসী কর্মকান্ড। | Behavior, Political |

Figure 4: Examplatory Sample for Comments & Category

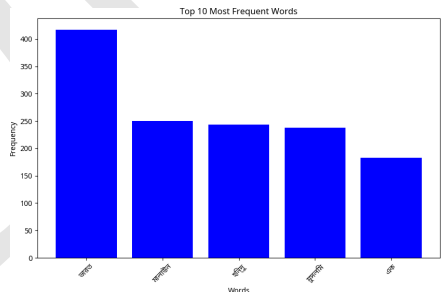

Figure 5: Most Frequently Occured Words

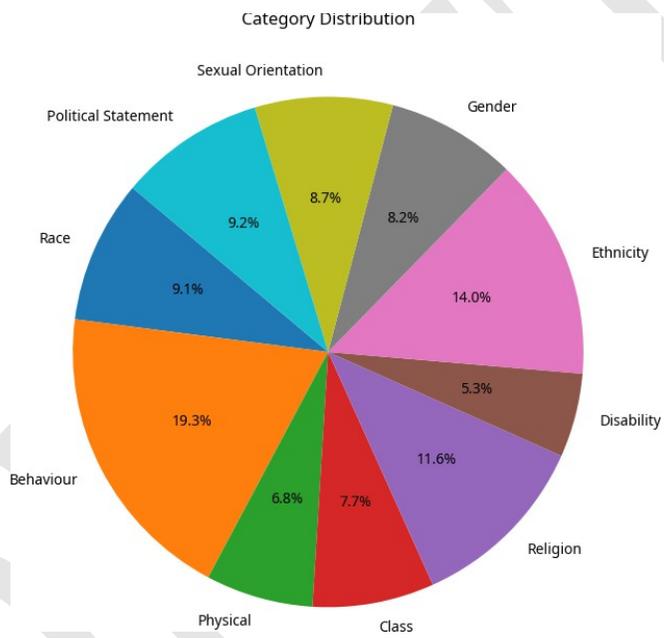

Figure 6: Category-wise Data Distribution

## 4 Experimental Process

The classification task was segmented into multiple phases to address the intricacies of the multi-label classification discussed in the preceding section. Such as

- Feature Selection
- Splitting Dataset
- Balancing Dataset
- Stop Word Handle
- Word Vectorization
- Model Selection
- Experiments
- Result Analysis

Figure 7 shows the steps involved in the experiments of hate speech classification.

### 4.1 Feature Selection

Feature selection is one of the most crucial tasks in any machine-learning approach. As this study focuses on a multi-label hate speech dataset, selecting appropriate features for classification is essential. From the previous section, Figure 6, it can be seen that disability has the lowest amount of data, which can cause problems during model training. To solve this problem, the Disability category has been discarded.

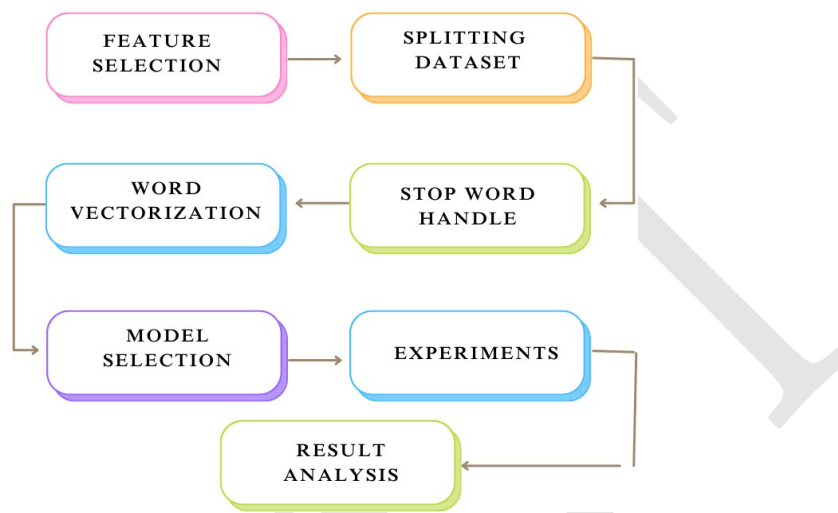

Figure 7: Experimental Workflow Diagram

## 4.2 Splitting Data

Before proceeding to the next step, we divided the data into three parts: Train, Test, and Validation.

Figure 8 shows the dataset split portions.

## 4.3 Stop Word Handle

Stop word refers to words that don't retain valuable patterns in documents. This study uses common stop words in the Bangla language and removes them before the word vectorization process.

Figure 9 shows some predefined stopwords used in the Bangla language.

## 4.4 Word Vectorization

Word vectorization is a method for transforming words into numeric values, which can be used in machine learning tasks. Vectorization consists of several steps.

**Data Split**

Train 80%  Validation 10%  Test 10%

Figure 8: Adopted Data Splitting Mechanism

Figure 9: Sample Stop words Distribution

- Tokenization
- Stemming
- Vocabulary Building
- Word transformation

This study uses a custom space tokenizer function and "Bangla Stemmer" to tokenize and stem Bangla words [Mahmud et al., 2014]. Then, it uses the scikit-learn, tf-idf vectorizer to finish the word vectorization process.

## 4.5 Model Selection

This study focuses on three widely used machine learning algorithms of three distinct categories from scikit-learn, namely

- Random Forest: Tree-Based ensemble model
- Support Vector Machine: Geometry or Boundry-based model
- Logistic Regression: Statistical Model

However, these models were not suited to handle multi-label classification tasks. That is why a wrapper function named MultiOutputClassifier was added so that the models could compare the multiple labels of the data instances.

### 4.5.1 Multi-Label Classification

MultiOutputClassifier from the scikit-learn library, designed to handle multi-output (or multi-label) classification problems. Let us consider a dataset with $n$ samples and $m$ features, represented as $X = \{x_1, x_2, \ldots, x_n\}$, and corresponding labels $Y = \{y_1, y_2, \ldots, y_k\}$, where each $y_j$ is a label for a specific class, and $k$ represents the number of possible labels (outputs).

The goal of multi-label classification is to learn a set of functions, one for each label, of the form:

$$f_j : X \to y_j, \quad j = 1, 2, \ldots, k$$

where $f_j$ denotes the classifier for the $j$-th label. The final prediction for an instance $X$ is a vector:

$$\hat{Y} = [\hat{y}_1, \hat{y}_2, \ldots, \hat{y}_k]$$

where each $\hat{y}_j$ is predicted independently by the classifier $f_j$.

### 4.5.2 Random Forest Classifier

The RandomForestClassifier is used as the base estimator in the multi-output setup. Random Forest is an ensemble method that constructs $t$ decision trees $h_1^{(j)}, h_2^{(j)}, \ldots, h_t^{(j)}$ for each label $y_j$, where each tree is trained on a bootstrapped (randomly sampled with replacement) subset of the data. The prediction for each label $y_j$ is given by the majority vote of the trees:

$$\hat{y}_j = \text{majority\_vote}(h_1^{(j)}(X), h_2^{(j)}(X), \ldots, h_t^{(j)}(X))$$

**Gini Impurity**

The decision trees within the random forest are constructed by splitting nodes based on minimizing the Gini impurity. For a dataset $D$, the Gini impurity is defined as:

$$Gini(D) = 1 - \sum_{i=1}^{C} p_i^2$$

where $p_i$ represents the proportion of class $i$ in the dataset $D$, and $C$ is the total number of classes. The splitting criterion selects the feature that minimizes the Gini impurity after the split.

### 4.5.3 Support Vector Classification

The Support Vector Classifier (SVC) model uses the sigmoid kernel to transform the input space non-linearly. The kernel function is defined as:

$$K(x_i, x_j) = \tanh(\gamma x_i^T x_j + r)$$

Where:

- $\gamma$ is the scaling parameter,
- $r$ is the kernel's intercept term.

The decision function for the $j$-th binary classifier is:

$$f_j(x) = \sum_{i=1}^{n} a_i^j y_i^j \tanh(\gamma x_i^T x + r) + b_j$$

Where $a_i^j$ are the Lagrange multipliers corresponding to the training samples, and $y_i^j$ is the label for the $i$-th sample.

**Decision Function Shape: One-vs-Rest (OvR)**

The decision_function_shape='ovr' parameter means that the decision function uses the One-vs-Rest (OvR) strategy for each label. This approach allows us to handle multi-label problems by fitting $L$ binary classifiers, where each classifier separates one label from all the others.

**Tie-Breaking Mechanism**

When break_ties=True, the model resolves ties that might occur when multiple labels have similar decision scores. The tie-breaking mechanism selects the label with the highest confidence score from the decision function:

$$\hat{y}_j = \arg\max_j f_j(x)$$

where $f_j(x)$ is the decision function score for the $j$-th label. This ensures that the classifier selects the most confident label for ambiguous predictions.

### 4.5.4 Logistic Regression

Logistic regression is a statistical method used for binary classification problems. It models the probability that a given input point belongs to a particular category. The logistic function (the sigmoid function) maps predicted values to probabilities.

The logistic function is defined as:

$$\sigma(z) = \frac{1}{1 + e^{-z}}$$

Where $z$ is a linear combination of the input features, expressed as:

$$z = \beta_0 + \beta_1 x_1 + \beta_2 x_2 + \ldots + \beta_n x_n$$

Here, $\beta_0$ is the intercept, and $\beta_1, \ldots, \beta_n$ are the coefficients for the features $x_1, x_2, \ldots, x_n$.

### 4.5.5 Final Model

Thus, the MultiOutputClassifier trains $k$ independent classifiers, one for each output label $y_j$. Each classifier collectively predicts labels using majority voting.

## 4.6 Evaluation Matrices

As the tasks involve multi-label classification, this study only focuses on the Macro-Average F1 Score. Macro-averaging is a technique for evaluating classifier performance in multi-class or multi-label settings. It involves independently calculating performance metrics—such as precision, recall, and F1 score—for each class and then averaging them across all classes. This approach ensures equal weighting for each class, which is particularly beneficial in scenarios with imbalanced class distributions.

The macro-averaged metrics are computed as follows:

$$\text{Macro-Precision} = \frac{1}{k} \sum_{j=1}^{k} \text{Precision}_j$$

$$\text{Macro-Recall} = \frac{1}{k} \sum_{j=1}^{k} \text{Recall}_j$$

$$\text{Macro-F1} = \frac{1}{k} \sum_{j=1}^{k} \text{F1}_j$$

Where $k$ is the number of classes.

Macro-averaging is particularly useful when the performance of minority classes is critical. It highlights the classifier's ability to perform well across all categories, avoiding bias towards majority classes.

## 5 Result Analysis

Figure 10 is a bar chart representing the Macro F1 Scores of selected three machine learning models: Logistic Regression, Support Vector Machine (SVM), and Random Forest. The F1 Score is a metric to evaluate a classification model's performance, especially when dealing with imbalanced datasets. In the bar chart, the Random Forest model performs best with an F1 score of 86.0%, followed by SVM with an F1 score of 81.04%. However, Logistic Regression somehow falls behind in the F1 scoring compared to the other two, with an F1 score of 60.0%. This concludes that the Random Forest model is more effective in handling the classification task of our dataset, which justifies our decision to choose it.

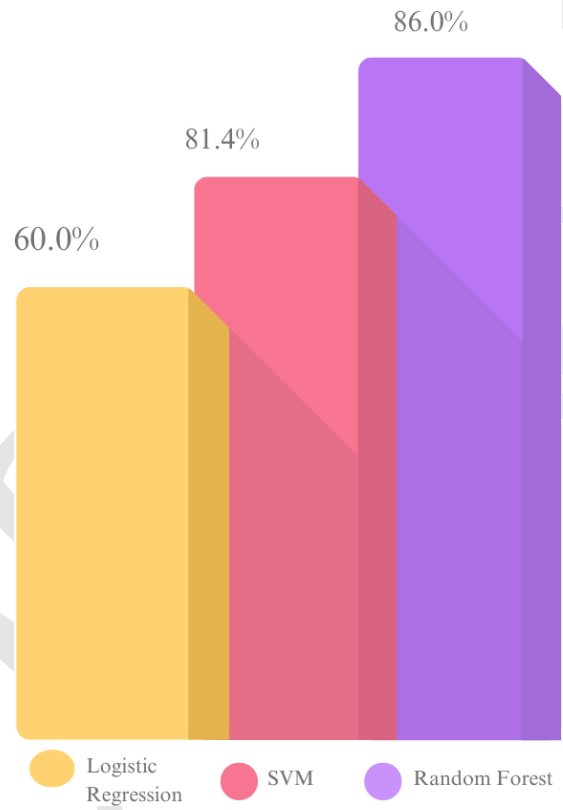

Figure 10: Comparative Landscape of F1 Score Across Algorithms

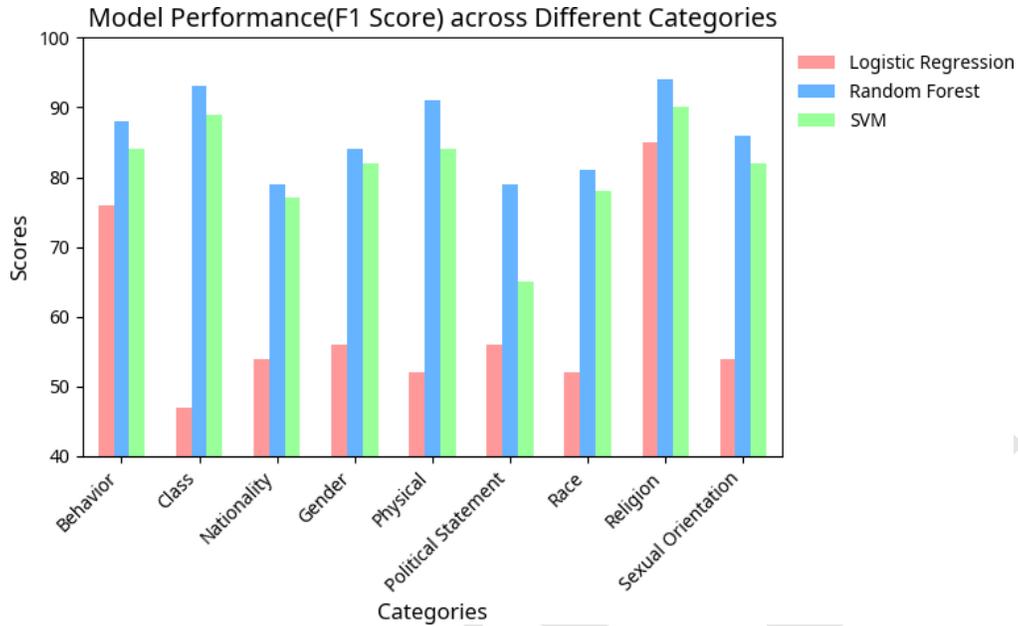

Figure 11: Comparative Landscape of F1 Score Across Categories

Figure 11 compares the performance of three well-known machine learning models, Logistic Regression, Random Forest, and SVM (Support Vector Machine), in terms of their F1 scores in the horizontal axis, there are nine distinct categories: Behavior, Class, Nationality, Gender, Physical, Political Statement, Race, Religion, and Sexual Orientation and F1 scores are plotted in the vertical axis, ranging from 40% to 100%. The Random Forest model consistently outperforms the other two models in most categories. If watched closely, the performance of the Random Forest model is specifically vital in categories like Religion, Race, and Physical, with F1 scores of approximately 95%, 93%, and 90%, respectively, indicating a near-optimal performance. Additionally, Random Forest scores above 80% in categories like Gender,

Behavior, class, and sexual orientation, demonstrating its strength in these areas. Furthermore, the F1 scores in Nationality and Political are above 70%, indicating moderate performance in these areas. The second model, the Support Vector Machine (SVM), performs relatively well. However, it often trails slightly behind Random Forest. However, it maintains strong scores across several categories like Gender, Physical, and Race. It is worth noticing that the SVM F1 scores range between 80% and 95%, demonstrating a robust predictive strength. However, its performance is comparatively weaker in categories like class and political statements, where its scores are notably lower, although they are still higher than those of logistic regression. Random Forest and SVM offer some acceptable performance, whereas Logistic Regression consistently lags behind both Random Forest and SVM. Its F1 scores generally range between 50% and 80%, with deficient performance in categories such as Class, Gender, and Political statement with scores around 50-60%. Even though Logistic Regression shows relatively better performance in categories such as behavior and race with scores closer to 75-80%, these results are still significantly lower than the scores achieved by Random Forest and SVM. In terms of overall condition, all three models show higher performance in categories such as Religion and Physical, which suggests that these categories are more easily predicted. Conversely, for categories like Class and Political Statement, the F1 scores are relatively lower in all three models, suggesting that these categories may present more complex challenges for machine learning models. In short, Random Forest consistently delivers the best performance across all categories compared to the rest of the models. SVM performs moderately well but with some slight inconsistency across different categories. However, Logistic Regression, by contrast, demonstrates the weakest overall performance among the three categories.

These results highlight strengths and limitations in handling different Random Forest and SVM categories. Additionally, Random Forest's consistently superior performance, particularly in categories like Religion, Race, and Physical Hate Speech, can be attributed to not only its ability to handle complex decision boundaries and interaction between features but also a balanced and well-structured dataset. Notably, despite being a traditional machine learning algorithm, Random Forest demonstrates competitive performance against more modern deep learning models like BERT and its variant on hate speech classification tasks. This holds not only for Bangla but also for similar low-resource languages, such as Hindi and Arabic, which have fewer exclusive categories, as shown in studies by Mishara et al. and Arshad et al. In these studies, Mishra's multi-task BERT model achieved an F1 score of 85%, while Arshad's BERT model reported an F1 score of 82% [Mishra et al., 2021, Arshad et al., 2023]. Additionally, these results reflect the performance observed on similar datasets for high-resource languages, such as English, across both deep learning and traditional machine learning approaches. For instance, Abro's study reported an SVM model with an F1 score of 79%, while Mercan's research achieved F1 scores of 87% with BERT and 84% with SVM. In contrast, Warner's earlier work demonstrated a considerably lower SVM performance, with an F1 score of 63%, highlighting the variability in effectiveness across different models and datasets [Abro et al., 2020, Mercan et al., 2021, Warner and Hirschberg, 2012].

These findings underscore the effectiveness of traditional machine learning approaches like Random Forest in handling hate speech classification tasks, particularly for low-resource languages such as Bangla, Hindi, and Arabic. The model's strong performance can be attributed to its capability to manage complex decision boundaries and the advantages of using a balanced dataset, enabling it to compete effectively with more modern deep learning methods like BERT. The variation in the performance of SVM and BERT across different studies and languages further illustrates the importance of aligning model selection with the characteristics of the dataset and classification task. This highlights the nuanced trade-offs in leveraging traditional and advanced hate speech detection techniques across diverse linguistic contexts.

# 6 Conclusion

This study effectively fills the gaps in Bangla hate speech detection by developing a comprehensive multi-label dataset that covers various categories, including race, gender, politics, and more. Given the complexity and low-resource nature of the Bangla language, this dataset serves as a valuable resource for research. This dataset lays a foundation for creating machine learning models to combat hate speech on social media. By offering various annotated examples, algorithms can understand harmful language nuances and underlying motivations, thereby improving detection accuracy and contextual awareness. Its influence is not limited to technology; it also enables academic research in sociology, linguistics, and communication studies. This dataset provides critical insights that can catalyze social change and promote more respectful online interactions by elucidating the manifestation and dissemination of hate speech within the Bangla-speaking community.

Furthermore, the study offers an in-depth analysis, providing evidence-based insights into the performance of machine learning models for multi-label classification tasks. In addition, it validates the strategic preprocessing approach and demonstrates the effectiveness of the machine learning models in handling its complexities. This paper highlights the importance of enhancing hate speech detection models for low-resource languages to create a better, safer, and more civilized online community. Additionally, the datasets and findings provided in this paper for hate speech detection in Bangla can serve as a great foundational resource for further advanced research and studies in this field. Therefore, this paper contributes to the vast field of hate speech detection by including a properly well-managed dataset that focuses on and highlights a low-resource language, Bangla, and paves the way for the Bangla language to have a broader impact on the global digital landscape. This study doesn't focus on deep learning algorithms for classification tasks.

Despite the progress made, this study opens up several opportunities for enhancement. Expanding the dataset to include additional data points has the potential to improve model performance and generalizability. Furthermore, increasing the representation of currently under-represented categories would balance the class distribution, leading to better detection across all hate speech categories. Additionally, future exploration into deep learning techniques could offer improved accuracy and deeper insight compared to traditional machine learning models applied here.

Md. Redowan Mahmud, Mahbuba Afrin, Md. Abdur Razzaque, Ellis Miller, and Joel Iwashige. A rule based bengali stemmer. In *2014 International Conference on Advances in Computing, Communications and Informatics (ICACCI)*, pages 2750–2756, September 2014. doi: 10.1109/ICACCI.2014.6968484. URL https://ieeexplore.ieee.org/document/6968484.

Abdullah Al Maruf, Ahmad Jainul Abidin, Md Mahmudul Haque, Zakaria Masud Jiyad, Aditi Golder, Raaid Alubady, and Zeyar Aung. Hate speech detection in the bengali language: a comprehensive survey. *Journal of Big Data*, 11:1–53, 12 2024. ISSN 21961115. doi: 10.1186/S40537-024-00956-Z/FIGURES/17. URL https://journalofbigdata.springeropen.com/articles/10.1186/s40537-024-00956-z.

Vildan Mercan, Akhtar Jamil, Alaa Ali Hameed, Irfan Ahmed Magsi, Sibghatullah Bazai, and Syed Attique Shah. Hate speech and offensive language detection from social media. *2021 International Conference on Computing, Electronic and Electrical Engineering, ICE Cube 2021 - Proceedings*, 2021. doi: 10.1109/ICECUBE53880.2021.9628255.

Sudhanshu Mishra, Shivangi Prasad, and Shubhanshu Mishra. Exploring multi-task multi-lingual learning of transformer models for hate speech and offensive speech identification in social media. *SN Computer Science*, 2:1–19, 4 2021. ISSN 26618907. doi: 10.1007/S42979-021-00455-5/FIGURES/6. URL https://link.springer.com/article/10.1007/s42979-021-00455-5.

Sandip Modha, Thomas Mandl, Prasenjit Majumder, and Daksh Patel. Tracking hate in social media: Evaluation, challenges and approaches. *SN Computer Science*, 1:1–16, 3 2020. ISSN 26618907. doi: 10.1007/S42979-020-0082-0/METRICS. URL https://link.springer.com/article/10.1007/s42979-020-0082-0.

Megan A. Moreno, Aubrey D. Gower, Heather Brittain, and Tracy Vaillancourt. Applying Natural Language Processing to Evaluate News Media Coverage of Bullying and Cyberbullying. *Prevention Science*, 20(8):1274–1283, November 2019. ISSN 1573-6695. doi: 10.1007/s11121-019-01029-x. URL https://doi.org/10.1007/s11121-019-01029-x.

M. F. Mridha, Md Anwar Hussen Wadud, Md Abdul Hamid, Muhammad Mostafa Monowar, M. Abdullah-Al-Wadud, and Atif Alamri. L-boost:
31